\documentclass[journal]{ieeetran}

\usepackage{graphics}           
\usepackage{times}              
\usepackage{amsmath}            
\usepackage{amssymb}            
\usepackage{graphicx}
\usepackage{algorithm}
\usepackage[noend]{algpseudocode}
\usepackage{booktabs}
\usepackage{color}
\usepackage{subfigure}
\usepackage{multirow}
\usepackage{setspace}
\definecolor{instructioncolor}{rgb}{.5,.5,.5}
\usepackage{threeparttable}
\usepackage{xfrac}
\usepackage{makecell}
\usepackage{bbding}
\usepackage{balance}
\makeatletter
\usepackage[colorlinks=true,
            linkcolor=blue, 
            anchorcolor=blue, 
            citecolor=green
            ]{hyperref}

\renewcommand{\maketag@@@}[1]{\hbox{\m@th\normalsize\normalfont#1}}%
\makeatother

\usepackage[font=small]{caption}

\def\secref#1{Sec.~\ref{#1}}
\def\figref#1{Fig.~\ref{#1}}
\def\tabref#1{Tab.~\ref{#1}}
\def\eqref#1{Eq.~(\ref{#1})}

\makeatletter
\usepackage{xspace}
\DeclareRobustCommand\onedot{\futurelet\@let@token\@onedot}
\def\@onedot{\ifx\@let@token.\else.\null\fi\xspace}

\def\etal{{et al}\onedot}
\makeatother

\def\etalcite#1{\etal~\cite{#1}}

\usepackage{array}
\newcolumntype{L}[1]{>{\raggedright\let\newline\\\arraybackslash\hspace{0pt}}m{#1}}
\newcolumntype{C}[1]{>{\centering\let\newline\\\arraybackslash\hspace{0pt}}m{#1}}
\newcolumntype{R}[1]{>{\raggedleft\let\newline\\\arraybackslash\hspace{0pt}}m{#1}}

\newcommand{\wang}[1]{{\textcolor{black}{#1}}}

\newcommand{\RR}{\mathbb{R}}

\renewcommand{\b}[1]{\mbox{\boldmath$#1$}}

\newcommand{\m}[1]{{\mbox{{\sffamily\slshape{#1\/}}}}}

\newcommand{\tr}[0]{\sf T}              

\newcommand{\bF}{\b F}

\newcommand{\bT}{\b T}

\newcommand{\bp}{\b p}

\newcommand{\mB}{\m B}

\newcommand{\mI}{\m I}

\newcommand{\mR}{\m R}
\newcommand{\mS}{\m S}

\usepackage{url}
\usepackage[numbers,sort&compress]{natbib}

\begin{document}

\title{SegNet4D: Efficient Instance-Aware 4D Semantic Segmentation for LiDAR Point Cloud}


\author{
        Neng Wang$^*$, \quad 
        Ruibin Guo$^*$, 
        \quad 
        Chenghao Shi, 
        \quad 
        Ziyue Wang, 
        \quad 
        Hui Zhang, \\
        Huimin Lu$^\dag$, 
        \quad 
        Zhiqiang Zheng, 
        \quad 
        Xieyuanli Chen$^\dag$
 
  \thanks{All authors are with the College of Intelligence Science and Technology, National University of Defense Technology, Changsha, China.}%
  \thanks{$^*$ indicates the equal contribution.}
  \thanks{$^\dag$ Joint corresponding authors: \{lhmnew, xieyuanli.chen\}@nudt.edu.cn}
  \thanks{\wang{This work was supported in part by the National Science Foundation of China (Grant No. 62403478, U22A2059 and 62203460), Young Elite Scientists Sponsorship Program by CAST (No. 2023QNRC001), and  Major Project of Natural Science Foundation of Hunan Province (Grant No. 2021JC0004).}
}%
}

\maketitle

\begin{abstract}
  4D LiDAR semantic segmentation, also referred to as multi-scan semantic segmentation, plays a crucial role in enhancing the environmental understanding capabilities of autonomous vehicles or robots. It classifies the semantic category of each LiDAR measurement point and detects whether it is dynamic, a critical ability for tasks like obstacle avoidance and autonomous navigation. Existing approaches often rely on computationally heavy 4D convolutions or recursive networks, which result in poor real-time performance, making them unsuitable for online robotics and autonomous driving applications.
  In this paper, we introduce SegNet4D, a novel real-time 4D semantic segmentation network offering both efficiency and strong semantic understanding. SegNet4D addresses 4D segmentation as two tasks: single-scan semantic segmentation and moving object segmentation, each tackled by a separate network head. Both results are combined in a motion-semantic fusion module to achieve comprehensive 4D segmentation.
  Additionally, instance information is extracted from the current scan and exploited for instance-wise segmentation consistency. Our approach surpasses state-of-the-art in both multi-scan semantic segmentation and moving object segmentation while offering greater efficiency, enabling real-time operation.
  Besides, its effectiveness and efficiency have also been validated on a real-world unmanned ground platform.
  Our code will be released at \url{https://github.com/nubot-nudt/SegNet4D}.
\end{abstract}

\section{Introduction}
\label{sec:intro}
Light Detection and Ranging (LiDAR), employing time of flight (ToF) measurement technology~\cite{Xin2023tim}, enables accurately capturing environmental geometry details. 
Known for its resilience to illumination changes, wide field of views, and accurate measurement, LiDAR have become an important instrument for autonomous vehicles and robots.
As a result, LiDAR-based semantic perception has garnered extensive research interest in recent years.
Semantic segmentation using 3D LiDAR data aims to assign a specific semantic class label to each point in the acquired  environmental measurements, serving as a fundamental task in 3D perception and scene understanding.
It plays a critical role in enhancing downstream tasks such as point cloud registration~\cite{shi2023tits} and simultaneous localization and mapping (SLAM)~\cite{chen2019iros}.

In practical applications, however, single-scan 3D LiDAR semantic segmentation methods~\cite{zhou2022tpami,Wang2023tim,thomas2019iccv,li2022ral} are limited because they lack motion information, which is critical for obstacle avoidance and path planning in autonomous robotic and driving applications. 
To address this, 4D semantic understanding is necessary, as it categorizes each LiDAR measurement while identifying its dynamic attributes~\cite{shi2020cvpr,Duerr20203dv,liu2023cvpr}.
This task typically relies on sequential LiDAR scans to capture motion information.
In this regard, some methods directly stack historical LiDAR scans into a single pointcloud and feed it into a single-scan-based network for achieving multi-scan segmentation~\cite{zhou2022tpami,thomas2019iccv,behley2019iccv}, which leads to suboptimal performance due to the lack of temporal information association.
Other approaches use 4D convolutional neural networks~\cite{choy2019cvpr} or recursive networks~\cite{Duerr20203dv,Schutt2022icra} to extract motion features from input sequential LiDAR scans.
However, these approaches impose heavy computational burdens and make real-time operations challenging.
Moreover, existing methods commonly treat multi-scan semantic segmentation in an end-to-end fashion, directly predicting semantic labels for all categories, including those moving and static classes.
This may lead to limited performance in recognizing dynamic objects because moving points are often fewer than static background points in most datasets.

To address these challenges, building upon our previous work InsMOS~\cite{wang2023iros}, we propose an extended framework for 4D semantic segmentation.
Our key idea is to tackle 4D semantic segmentation by dividing it into two subtasks: single-scan semantic segmentation (SSS) and moving object segmentation (MOS). We argue that semantic classes and motion characteristics represent distinct aspects of objects. By addressing them separately and then fusing their outputs, our method achieves a more comprehensive understanding and superior performance compared to end-to-end approaches.
To enhance efficiency, our framework converts the sequential LiDAR scans into Bird's Eye View (BEV) images and extracts motion features by calculating BEV residuals, significantly reducing computational cost compared to existing fashion using 4D convolutions.
Furthermore,
we find instance information~\cite{chen2022ral,wang2023iros} is crucial for the LiDAR segmentation tasks to prevent over-segmentation. However, existing methods struggle to integrate this information effectively.
To address this, we incorporate instance consistency from the current scan into the prediction pipeline at both feature and point levels, enabling instance-aware segmentation.
Finally, we design a novel module to fuse point-wise semantic predictions and motion states to achieve accurate 4D semantic segmentation online.
We extensively evaluate SegNet4D on mainstream datasets, comparing its performance with state-of-the-art (SOTA) methods. The results confirm that our approach surpasses SOTA in both multi-scan semantic segmentation and MOS while supporting real-time operation. Additionally, we validate its practical utility on a real unmanned ground platform, demonstrating both its effectiveness and efficiency for autonomous applications.

In summary, the contributions of this work are threefold:
\begin{itemize}
     \item We propose a novel 4D semantic segmentation framework for LiDAR measurements, which decomposes the task into two subtasks: Single-Scan Semantic Segmentation (SSS) and Moving Object Segmentation (MOS). The results are then combined using a motion-semantic fusion module, achieving SOTA performance in both 4D semantic segmentation and MOS. 
     \item We propose a novel instance-aware backbone that leverages instance information at both the feature and point levels, enhancing performance in both MOS and semantic segmentation tasks. Additionally, we introduce an automated method to generate instance labels directly from semantic annotations, eliminating the need for additional manual labeling.
     \item We propose an efficient 4D semantic segmentation network that is significantly faster than existing methods and  supports real-time operation. Furthermore, we successfully integrate the network into a real platform, demonstrating its practical utility for autonomous applications.
\end{itemize}

This article is an extension of our previous conference paper, i.e., InsMOS~\cite{wang2023iros}, which proposed to add instance information for MOS. 
Compared to~\cite{wang2023iros}, this article extends it in four critical aspects:
1) a fully 4D semantic segmentation method that predicts both the motion states and semantic categories.
2) a novel module for integrating point-wise motion states and semantic category predictions.
3) a more efficient fashion for encoding motion features, enabling this method to operate in real-time.
4) a more extensive experimental evaluation on multiple datasets and a real-world unmanned ground platform, demonstrating this method's superior performance and practical utility.

\section{Related Work}
\label{sec:related}

\subsection{Moving Object Segmentation.}    
LiDAR-based MOS refers to accurately identifying motion state for each measurement point. Existing methods can be divided into two groups, projection-based and non-projection-based fashion.

The projection-based method typically converts sequential point clouds into image representations, such as range images ~\cite{chen2021ral,kim2022ral,sun2022iros,cheng2024icra} or BEV images ~\cite{mohapatra2021arxiv,zhou2023ral}, and then compute the residuals of sequential images to extract motion information in the scene. 
Chen~\etalcite{chen2021ral} first released the MOS datasets and benchmark based on the SemanticKITTI dataset~\cite{behley2019iccv}, and proposed a deep neural network to learn motion cues from sequential residual range images for online MOS.
Subsequently, some work~\cite{sun2022iros,kim2022ral,cheng2024icra} attempt to enhance the network through encodes the appearance features and temporal motion features separately, and then fuses them with a multi-scale motion-guided attention module.
Unlike range images, BEV images offer a top-down perspective, allowing a more intuitive observation of object movement. 
Based on this, Mohapatra~\etalcite{mohapatra2021arxiv} designed a lightweight network architecture for extracting motion information from BEV residual images and successfully achieved real-time operation in an embedded platform. 
However, this method can only perform pixel-wise segmentation in the BEV space and has relatively low accuracy.
Recently, Zhou~\etalcite{zhou2023ral} projected 3D point clouds into the polar coordinate BEV space and designed a dual-branch network structure similar to~\cite{sun2022iros} to enhance the MOS performance.

Different from the projection-based methods, the non-projection methods directly extract motion information from sequential 3D point clouds. 
Mersch~\etalcite{mersch2022ral} utilized the Minkowski engine~\cite{choy2019cvpr} to construct a sparse 4D convolutional network.
This network is designed to extract spatio-temporal features from the input 4D point clouds and predict point-wise moving labels.
They also proposed a receding horizon strategy that integrates multiple observations to refine the network’s predictions.
Kreutz~\etalcite{kreutz2023wacv} proposed an unsupervised MOS method based on \cite{mersch2022ral}, which learns spatio-temporal occupancy changes in the local neighborhood of point cloud videos. However, it is only applicable to stationary LiDAR.
Due to the lack of instance-level perception, existing methods often partially segment moving objects.
Therefore, we attempt to introduce instance information into the network, enabling it to achieve complete segmentation of moving objects.

\vspace{-0.1cm}
\subsection{4D LiDAR Semantic Segmentation.} 
Unlike the MOS task, 4D LiDAR semantic segmentation not only needs to capture temporal information to predict point-wise motion state but also assign a semantic category label to each point, including moving and static points.
To address this task, some methods \cite{zhou2022tpami,thomas2019iccv,behley2019iccv} try to fuse sequential LiDAR scans into a single point cloud and then utilize a 3D semantic segmentation network to perform 4D semantic segmentation directly.
However, due to the lack of temporal perception capability in 3D networks, these methods often struggle with point-wise motion state recognition. 
Moreover, the fused single pointcloud contains a substantial amount of data, greatly increasing the network's computation time and making real-time operation challenging.

Another category of method builds temporal-aware networks to tackle multi-scan semantic segmentation tasks.
SpSequenceNet~\cite{shi2020cvpr} employs two modules based on 3D convolutions, a cross-frame global attention module and a cross-frame local interpolation module, to extract spatio-temporal information from input sequential LiDAR scans.
TemporalLidarSeg~\cite{Duerr20203dv} recursively extracts features from the sequential range image and utilizes a temporal memory alignment strategy to align the features of adjacent frames.
Similarly, MemorySeg~\cite{Li2023iccv} also employs a features alignment strategy and operates in 3D space, which enhances point-level and voxel-level temporal association.
Recently, Liu~\etalcite{liu2023cvpr} proposed a module, MarS3D, which converts the sequential point clouds into BEV images and then utilizes an image network to extract motion features. 
This module can provide temporal awareness for existing 3D semantic segmentation networks.
Besides, Chen~\etalcite{Chen2023iccv} proposed a sparse voxel-adjacent query network for processing temporal information,  improving 4D semantic segmentation performance significantly.

The above-mentioned approaches require extracting temporally associated features from the sequential point clouds, resulting in time-consuming network operation.
Although TemporalLidarSeg~\cite{Duerr20203dv} and MarS3D~\cite{liu2023cvpr} also utilize a projection-based fashion, they still rely on complex image processing networks to capture temporal features. 
In contrast, our method streamlines the process by directly incorporating pre-extracted motion residuals into the network as additional feature channels for network learning, which avoids intentionally extracting temporally associated features, thereby significantly boosting efficiency.
Finally, by leveraging two separate output heads to supervise MOS and SSS explicitly and fusing them for 4D semantic segmentation, our method maintains superior segmentation accuracy.

\begin{figure}[t]
	\centering
	\includegraphics[width=1\linewidth]{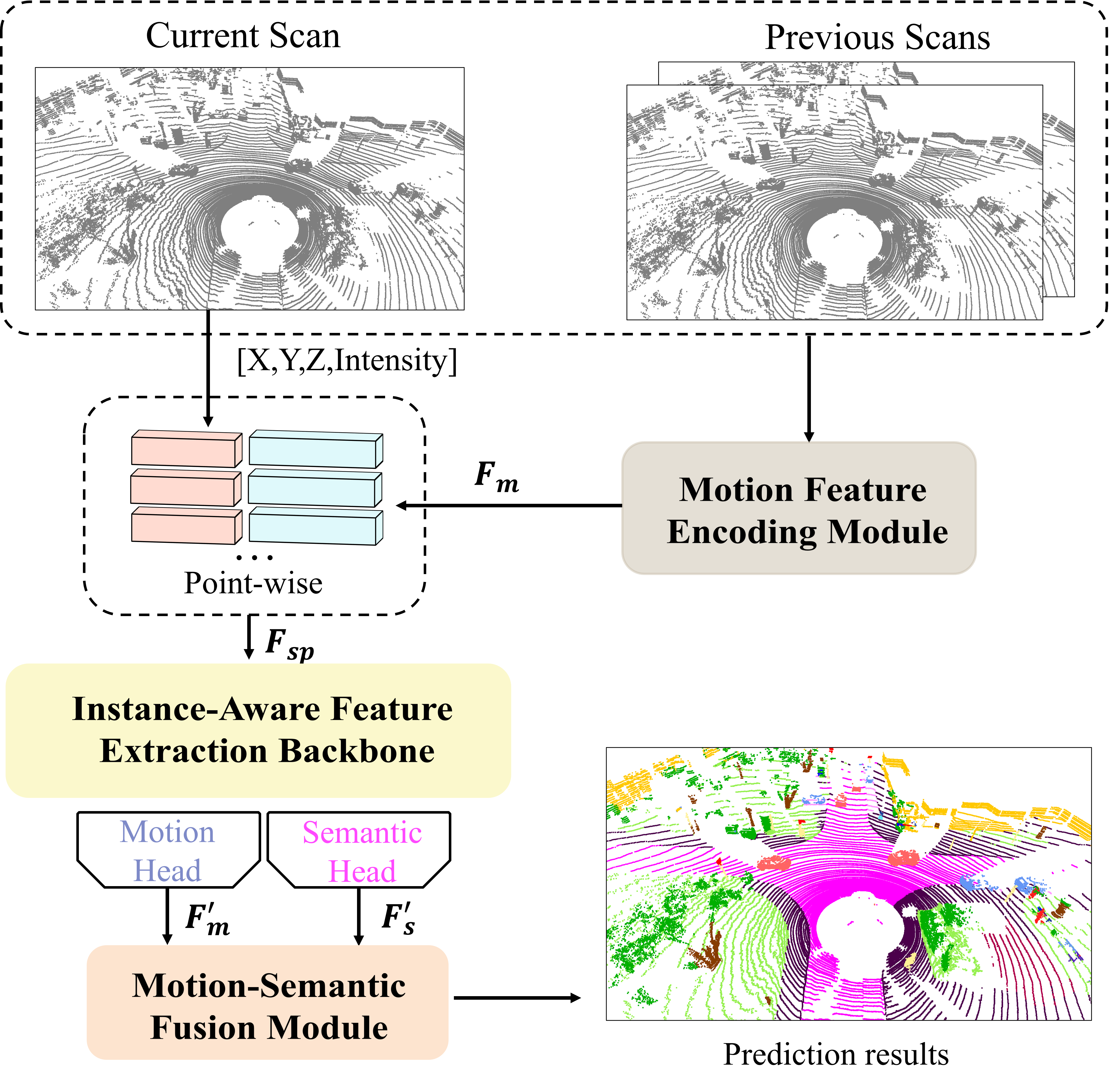}
	\caption{The proposed framework of SegNet4D. 
        We first utilize the Motion Features Encoding Module to extract motion features from the sequential LiDAR scans.
        Following this, the motion features are concatenated with the spatial features of the current scan and fed into the Instance-Aware Feature Extraction Backbone.
        Then, two separate heads are applied: a motion head for predicting moving states, and a semantic head for predicting semantic category.
		Finally, the Motion-Semantic Fusion Module integrates the motion and semantic features to achieve 4D semantic segmentation.
	}
	\label{fig:pipeline}
    \vspace{-0.3cm}
\end{figure}

\section{Methodology}
The framework of our proposed method is depicted in~\figref{fig:pipeline}. SegNet4D consists of four main components: Motion Feature Encoding Module (\secref{sec:MFE}), 
Instance-Aware Feature Extraction Backbone (\secref{sec:FEB}),
two separate output heads (\secref{sec:two_head}), and Motion-Semantic Fusion Module (MSFM,~\secref{sec:SMF}). 
We specifically introduce each module in the following section.

\subsection{Motion Feature Encoding Module}
\label{sec:MFE}
\begin{figure}[t]
	\centering
	\includegraphics[width=1\linewidth]{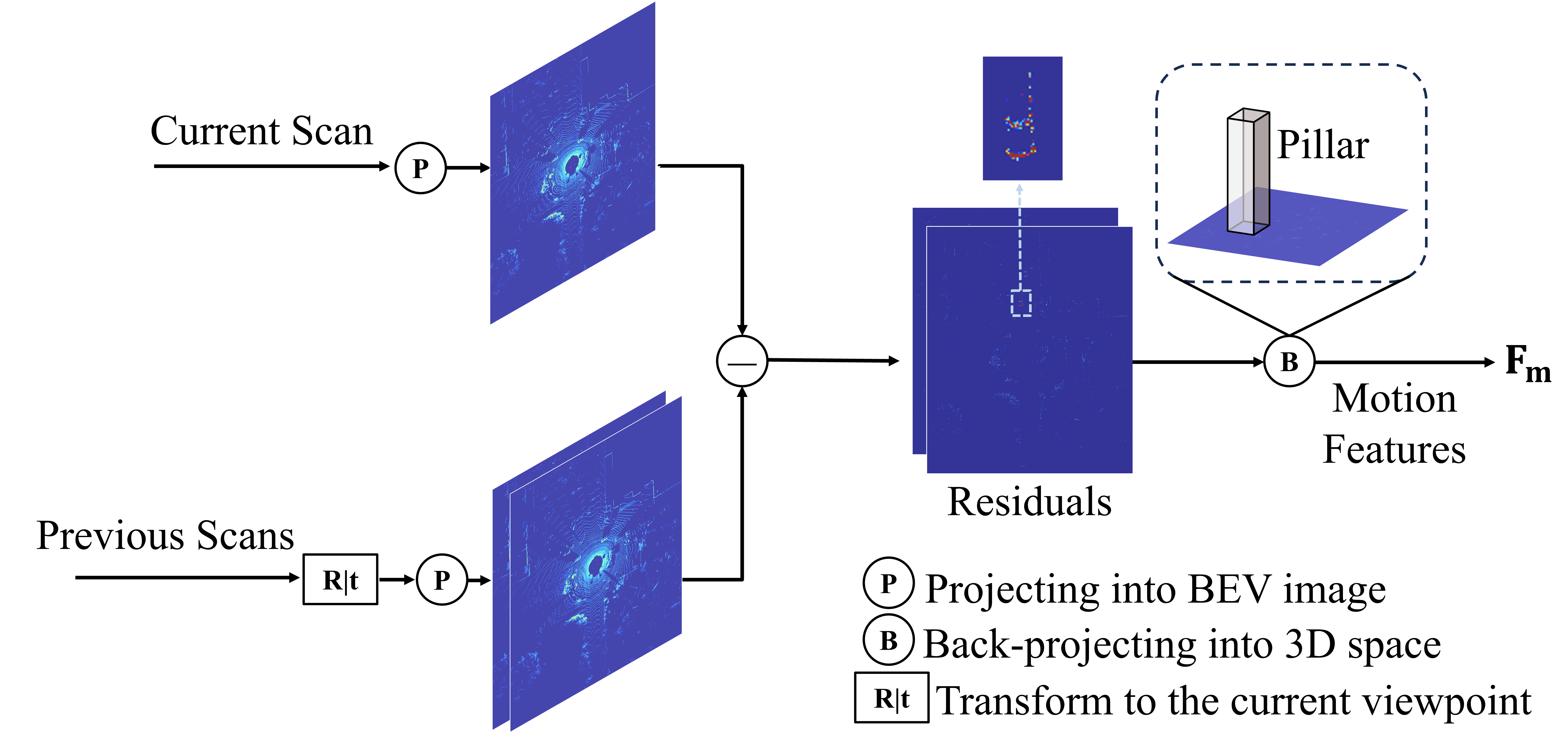}
	\caption{The process for motion features encoding. We mainly calculate the residuals for the sequential BEV images and back-project them into the 3D space as the motion features. 
	}
	\label{fig:motion_module}
 \vspace{-0.3cm}
\end{figure}

4D semantic segmentation not only predicts the semantic category for each point measured by LiDAR but also identifies its motion state. Therefore, it is necessary to extract motion features from the sequential point clouds. 
Existing methods mainly utilize 4D convolution~\cite{mersch2022ral,kreutz2023wacv} to obtain motion cues,
which poses a significant computational workload.
To improve the real-time performance, we adopt the BEV image representation for fast motion feature encoding, as shown in~\figref{fig:motion_module}, which avoids computational expense processing on extensive unstructured point clouds data. 
The encoding process is divided into three steps as follows.

\textbf{Point Cloud Alignment.}
To compensate for the ego-motion of LiDAR, spatial alignment is conducted initially on the input sequential point clouds.
Specifically,
given the current point cloud $\mS_{0}=\{\bp_{i} \in \RR^{4}\}^{M}_{i=1}$ consisting of $M$ points represented in homogeneous coordinates as $\bp_{i}=[x_{i},y_{i},z_{i},1]^{\tr}$, along with the past $N-1$ consecutive point clouds $\mS_{1},\mS_{2},...,\mS_{N-1}$ with their relative transformations $\bT_{1}^{0},\bT_{2}^{1},...,\bT_{N-1}^{N-2}$, we transform the past $N-1$ consecutive point clouds into the current viewpoint by
\begin{equation}
\mS_{j\rightarrow0}=\{\bp_{i}'=\bT_{j}^{0}\bp_{i}|\bp_{i}\in \mS_{j}\},~\bT_{j}^{0}\hspace{-1mm}=\hspace{-1mm}\prod_{k=0}^{j-1}\bT_{j-k}^{j-k-1}.
\end{equation} 
In practical applications, the relative transformations can be easily obtained through the existing LiDAR odometry approach~\cite{shi2023tits,chen2019iros}. 
We use the poses estimated by SuMa~\cite{chen2019iros}.

\textbf{BEV Projection.}
After alignment, we project the aligned point clouds into single-channel BEV images.
For each point $\bp_{j}'=(x_{j}',y_{j}',z_{j}') \in \mS_{j\rightarrow0}$, we first restrict it within $x_j' \in [X_{min},X_{max}]$, $y_j' \in [Y_{min},Y_{max}]$, $z_j' \in [Z_{min},Z_{max}]$ and then convert it into the pillar space, given by
\begin{equation}
\left\{
\begin{array}{l}
\mI_{(u,v),j}=\{z_{j}'|z_{j}'\in \bp_{j}'\}, \\
    u=\lfloor\frac{x_{j}' - X_{min}}{g}\rfloor, \\
    v=\lfloor\frac{y_{j}' - Y_{min}}{g}\rfloor,
\end{array}
\right.
\end{equation} 
where $\mI_{(u,v),j}$ stores a set of point's height in the pillar located at $(u,v)$, and  $g$ denotes grid resolution.

Following~\cite{zhou2023ral}, we project the $\mI_{j}$ into a single-channel BEV image $\mB_j$ of size $H \times W$. For each pixel value $\mB_{(u,v),j}$, we use the difference between the maximum and minimum height within the same pillar, calculating as:
\begin{equation}
\mB_{(u,v),j}=\text{Max}\{\mI_{(u,v),j}\}-\text{Min}\{\mI_{(u,v),j}\}
\end{equation} 

\textbf{Motion Features Encoding.} 
We take the BEV residuals $\mR \in \RR^{H \times W \times (N-1)}$ between $\mB_0$ and $\mB_{1},...,\mB_{N-1}$ as the motion features in the BEV space, calculated by
\begin{equation}
\mR_{(u,v),j\rightarrow0}=\mB_{(u,v),0}-\mB_{(u,v),j},j\in{1,...,N-1},
\end{equation}
where $\mR_{(u,v)}$ represents the residual value for pixel at $(u,v)$.
To obtain motion features for each point in the 3D space, we perform back-projection by assigning the BEV residual value to all points projected to the BEV pixel. 
Points within the same pillar will share the same residual value for each residual image.
Finally, we obtain point-wise motion features $\bF_{m} \in \RR^{M \times (N-1)}$.
We concatenate the $\bF_{m}$  and current scan's spatial features, i.e., $[$x, y, z, intensity$]$, generating a new features $\bF_{sp} \in \RR^{M \times (N+3)}$ as the input for subsequent backbone.

\begin{figure}[t]
	\centering
	\includegraphics[width=1\linewidth]{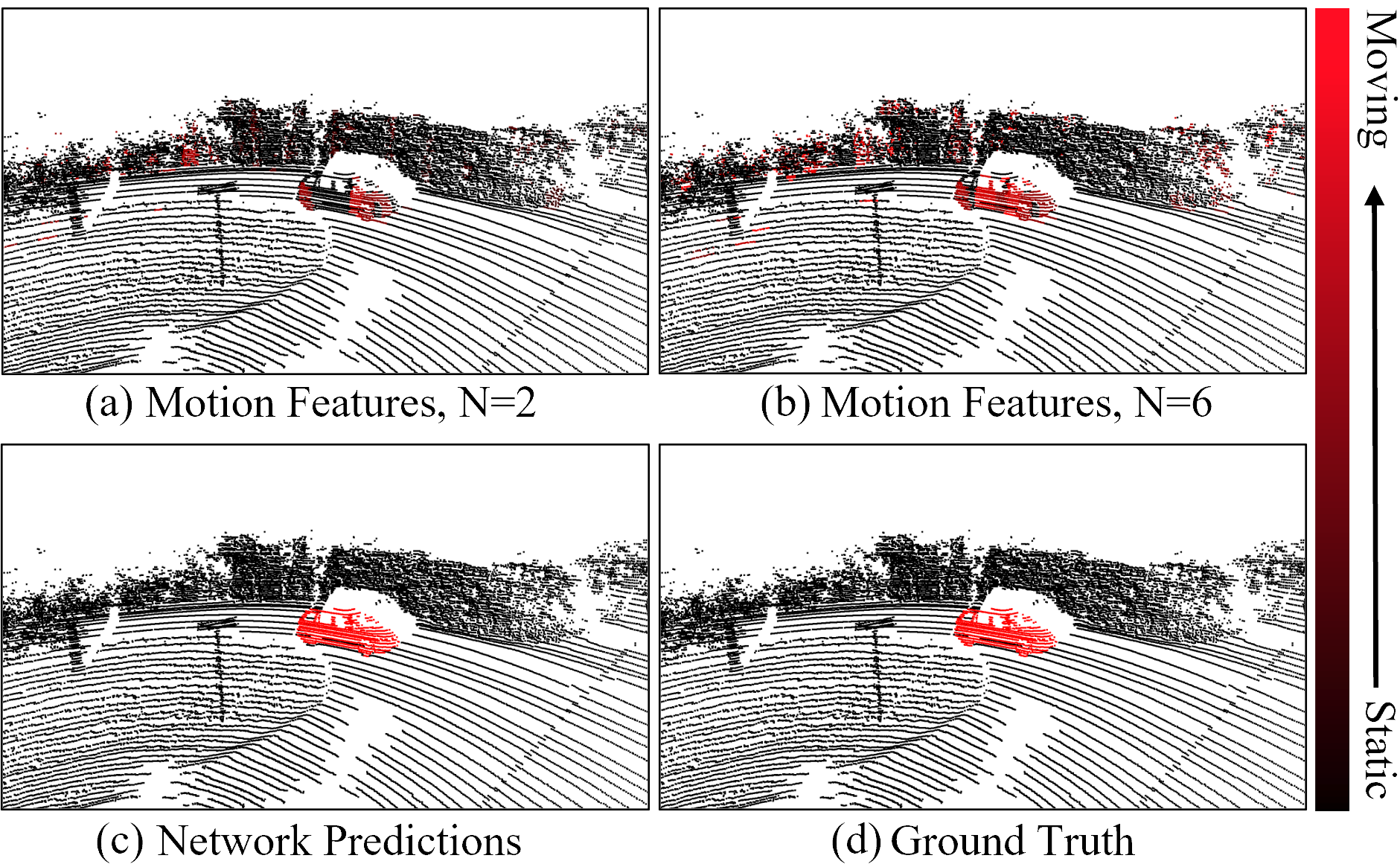}
	\caption{Motion features visualization. (a) and (b) represent the motion features obtained from the current and past $N$-th scan. We compare the features with the network's predictions as well as ground truth.
	}
	\label{fig:motion_features}
 
 \vspace{-0.3cm}

\end{figure}

The motion features visualization is presented in~\figref{fig:motion_features}. Our approach can extract initial motion cues for moving objects in the scene, and as the time interval extends, the motion features become increasingly prominent.
Inevitably, some points might be erroneously assigned motion characteristics because of changes in LiDAR viewpoint or inaccuracies in odometry estimation. However, through our network's subsequent learning process, which utilizes the scene's spatial features to distinguish movable from immovable objects and learning multi-channel motion features to identify true moving objects, 
these points still are correctly categorized, as shown in the~\figref{fig:motion_features}~(c).
This is also a significant difference from purely projection-based methods, as we still leverage valuable spatial features inherent in the original point cloud.

\subsection{Instance-Aware Feature Extraction Backbone}
\label{sec:FEB}
\begin{figure}[t]
	\centering
	\includegraphics[width=1\linewidth]{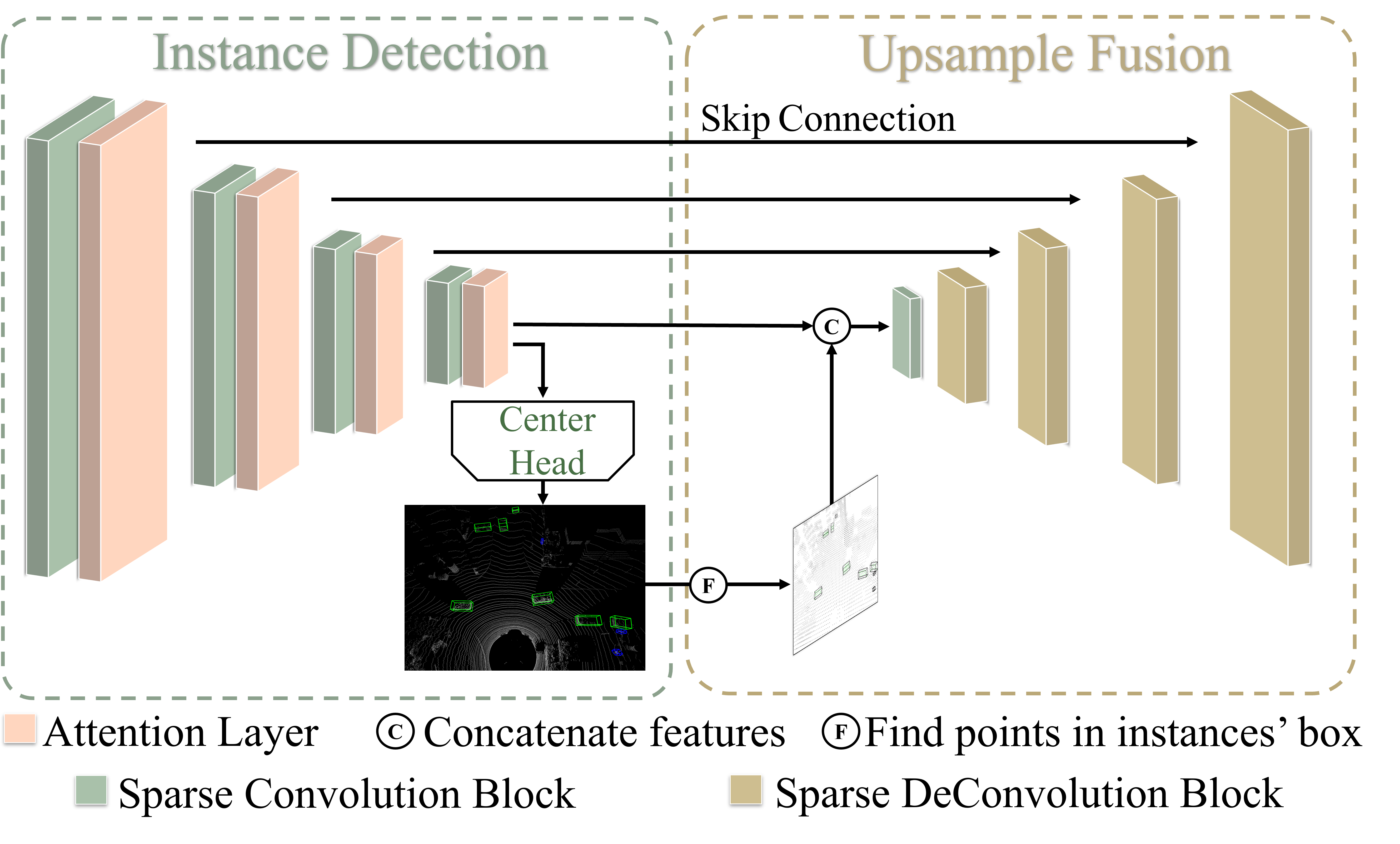}
	\caption{The architecture of the Instance-Aware Feature Extraction Backbone. The Instance Detection Module is utilized to extract features from the input point cloud and predict the instance bounding box. We then integrate such instance information into the Upsample Fusion Module to achieve instance-aware segmentation.
	}
	\label{fig:main_backbone}
 
 \vspace{-0.3cm}

\end{figure}

In semantic segmentation, over-segmentation is a common issue, where a single instance is incorrectly divided into multiple segments. 
Especially in the MOS task,  existing methods lack instance awareness, often erroneously dividing an instance into two different motion states.
To tackle such issues, we design the Instance-Aware Feature Extraction Backbone to extract instances information and incorporate them into our prediction pipeline.
As shown in~\figref{fig:main_backbone}. It consists of the Instance Detection Module and the Upsample Fusion Module.

\subsubsection{Instance Detection Module} 
\label{sec:IDM}
In this module,
We use 3D sparse convolution blocks~\cite{graham2018cvpr} to voxelize the current point cloud and then perform convolution only on non-empty voxels, greatly improving the processing speed.
Moreover, to capture more crucial features, we add a window self-attention~\cite{lai2023cvpr} layer after each sparse convolution block to improve the receptive field.
Our key finding is that instance information is crucial for LiDAR segmentation, as also shown in panoptic segmentation methods~\cite{Zhou2021cvpr,Rodrigo2023ral}. However, these methods typically require additional instance labels, which may not always be accessible. To address this, we propose a method to automatically generate instance bounding boxes using existing semantic annotations.
This process begins by generating each instance cluster from the semantic point cloud using the Euclidean clustering algorithm in PCL library~\cite{rusu2011icra}. Subsequently, these clusters are projected into the BEV space, and their orientation is determined using principal component analysis (PCA)~\cite{Hotelling1933AnalysisOA}. We then combine the height of the cluster to obtain the minimum bounding box and further refine it by L-Shape method~\cite{zhang2017iv}.
Finally, we utilize CenterHead~\cite{yin2021cvpr} trained with automatically generated instance labels, to predict instance bounding boxes online. These predicted boxes are then used to enhance feature extraction within our backbone, as described in the upsample fusion module.

\subsubsection{Upsample Fusion Module}
\label{sec:upsample}
To introduce instance consistency into feature-level fusion, we find and extract points within the instance bounding box as instance feature mask. Subsequently, we concatenate it with the point embeddings from the Instance Detection Module as input for the Upsample Fusion Module.
Unlike InsMOS~\cite{wang2023iros}, we integrate the instance features only once to enhance efficiency.
The Upsample Fusion Module serves the main purpose of integrating instance features into the prediction pipeline, as well as recovering the scale of the features hierarchically through 3D deconvolution blocks.
It is formed by a fully U-Net network structure together with the Instance Detection Module and maintains more details from different scale features through skip connection.

\subsection{Motion Head and Semantic Head}
\label{sec:two_head}
Existing 4D semantic segmentation methods usually predict all semantic class labels in an end-to-end manner, including those moving and static classes.
However, since static points typically outnumber moving ones in existing datasets, these approaches often result in suboptimal network performance in identifying moving classes.
So we employ two distinct heads for predicting moving labels and single-scan semantic labels separately. By explicitly supervising MOS, our method can maintain superior performance for moving object recognition.

To maintain the network's lightweight characteristics, these heads consist solely of a convolutional layer, a normalization layer, an activation function layer, and a linear layer, which are employed to further classify features with different attributes.
Finally, we can obtain point-wise motion features $\bF_{m}' \in \RR^{M \times 3}$ and semantic features $\bF_{s}' \in \RR^{M \times C}$.
By applying the softmax function, the network can output the motion labels and semantic labels of the current point cloud.
Note that $3$ represents three different motion classes, namely unlabeled, static, and moving, 
while $C$ denotes the number of semantic categories.
Each head is supervised with a specific loss function. Further details are provided in~\secref{sec:loss}.

\subsection{Motion-Semantic Fusion Module}
\label{sec:SMF}
\begin{figure}[t]
	\centering
	\includegraphics[width=1\linewidth]{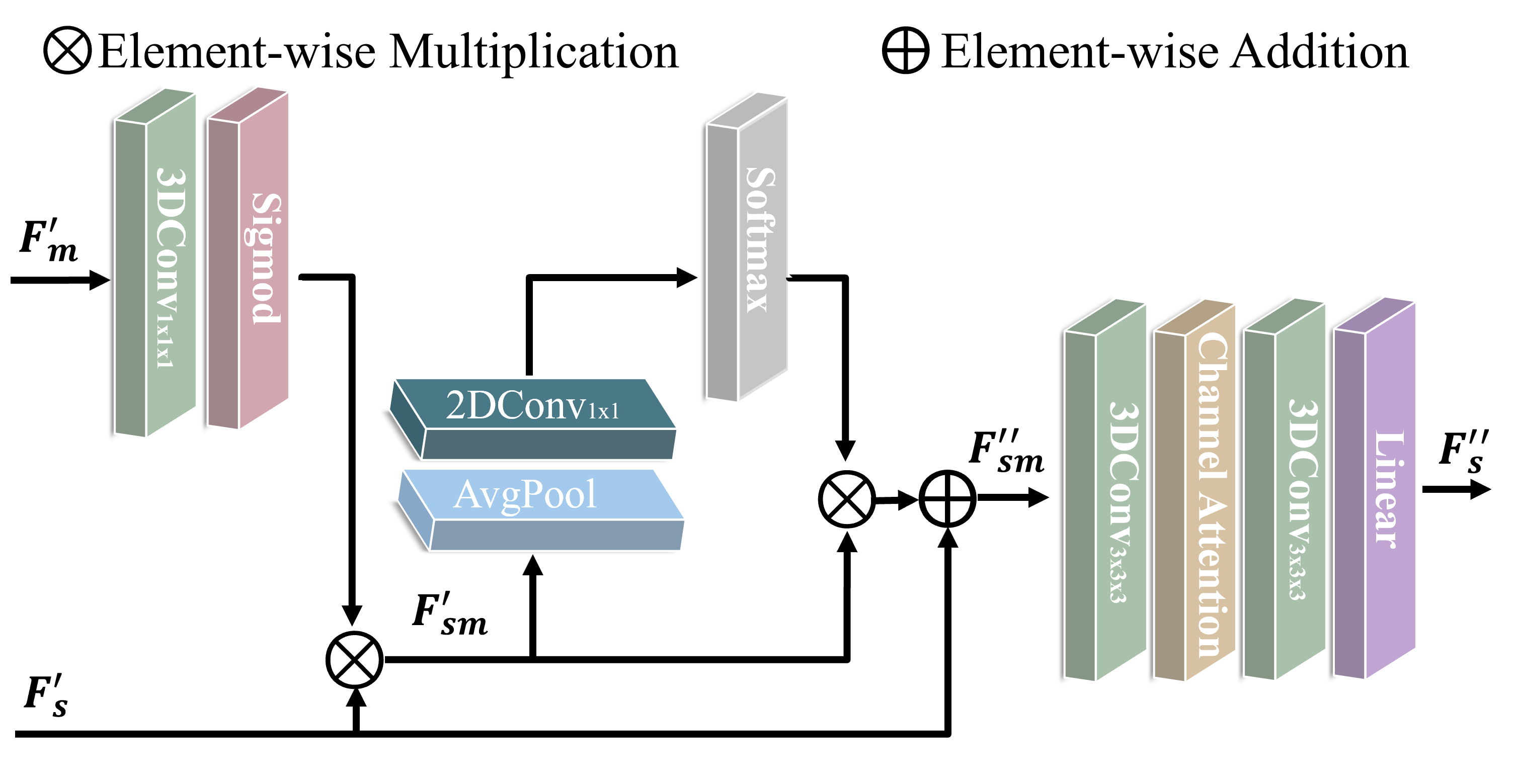}
	\caption{The architecture of Motion-Semantic Fusion Module. We mainly perform spatial attention and channel attention to fuse the motion features and the static semantic features.
	}
	\label{fig:smf}
 
 \vspace{-0.3cm}
\end{figure}

After obtaining motion labels and single-scan semantic labels,  a straightforward approach to achieve 4D semantic segmentation is to fuse them by checking the motion states on each semantic point. However, this fashion may yield non-smooth point segmentation. Moreover, incorrect motion predictions directly lead to false 4D semantic segmentation results.
We hence propose a motion-semantic fusion module to further integrate motion features $\bF_{m}'$ and static semantic features $\bF_{s}'$ for achieving motion-guided 4D semantic segmentation.
Details of this module are illustrated in~\figref{fig:smf}.

We build our Motion-Semantic Fusion Module upon recently advanced motion-guided attention module~\cite{lih2019iccv} in the field of image processing.
We extend it into 3D space and mainly utilize the 3D submanifold sparse convolution~\cite{graham2018cvpr} as the backbone for performing spatial attention between the $\bF_{s}'$ with $\bF_{m}'$ as:
\begin{equation}
\bF_{sm}' = \bF_{s}' \otimes \text{Sigmoid}(\text{3DConv}_{1\times1\times1}(\bF_{m}')),
\end{equation}
where $\otimes$ represent element-wise multiplication, $\text{3DConv}_{1\times1\times1}\text{(·)}$ represent a 3D submanifold sparse convolution with $1\times1\times1$ kernel size, and $\bF_{sm}'\in \RR^{M\times D}$ are the fused motion-salient features, and $D$ denotes the number of feature channels.
We then perform channel attention for $\bF_{sm}'$ to strengthen the responses of key attributes.
Subsequently, it is element-wisely added by $\bF_{s}'$ because the static semantic features are equally crucial for the 4D semantic segmentation. The final $\bF_{sm}''$ is calculated as:
\begin{equation}
\bF_{sm}'' \hspace{-0.1cm} =\hspace{-0.1cm} \bF_{sm}' \otimes [\text{Softmax}(\text{2DConv}_{1\times1}(\text{AvgPool}(\bF_{sm}'))) \cdot D]+ \bF_{s}',
\end{equation}
where $\text{2DConv}_{1\times1}\text{(·)}$ and $\text{AvgPool(·)}$ denote a 2D convolution with $1\times1$ kernel size and average pooling operation, respectively.
2D convolution is mainly used to quickly calculate channel attention weights, which is different from using 3D convolution for spatial attention as mentioned earlier.

Finally, we further refine $\bF_{sm}''$ using 3D sparse convolutions and channel attention blocks~\cite{woo2018eccv} to generate the final 4D semantic segmentation predictions $\bF_{s}''\in \RR^{M \times C'}$, where $C'$ is the number of categories for 4D semantic segmentation.

\subsection{Moving Instance Refinement}
\label{sec:Refinement}
To further enhance the accuracy of MOS, we check again the point-wise predictions within the instance bounding box and propose an instance-aware post-processing algorithm for point-level refinement, combining in a bottom-up and top-down fashion.

This bottom-up aims to refine the results of the following second cases. 
Firstly, if many points within an instance are moving, then the instance is considered to be moving.
Secondly, if the scene contains many moving vehicles, it is considered as a highly dynamic scene, such as the highway.
Vehicles in this scene will be more easily classified as the moving class with a lower confidence threshold.
During the top-down step, when an instance is identified as moving, all points within the instance are determined as moving, which is natural when considering each instance as a rigid body.
Besides, because motion is a continuous process, we consider an instance to be moving only when classified as a moving class in multiple observations.
We provide a detailed algorithmic process in our conference paper~\cite{wang2023iros}.

\begin{table*}[t]
	\fontsize{6.5pt}{7pt}\selectfont

        \setlength{\tabcolsep}{2.5pt}
	\caption{Multi-scan semantic segmentation performance evaluation on the SemanticKITTI dataset. “mov." denotes moving category. The best results are in bold.}
	\centering
	\begin{tabular}{l|ccccccccccccccccccccccccc|p{0.65cm}<{\centering}|p{0.45cm}<{\centering}}
		\toprule
		Methods & \rotatebox{90}{car} &\rotatebox{90}{bicycle}&\rotatebox{90}{motorcycle} &\rotatebox{90}{truck} &\rotatebox{90}{other-vehicle} & \rotatebox{90}{person}& \rotatebox{90}{bicyclist}& \rotatebox{90}{motorcyclist}&\rotatebox{90}{road} & \rotatebox{90}{parking}& \rotatebox{90}{sidewalk}&\rotatebox{90}{other-ground} & \rotatebox{90}{building}& \rotatebox{90}{fence}& \rotatebox{90}{vegetation}& \rotatebox{90}{trunk}& \rotatebox{90}{terrain}& \rotatebox{90}{pole}&\rotatebox{90}{traffic sign}& \rotatebox{90}{mov. car}& \rotatebox{90}{mov. truck}& \rotatebox{90}{mov. other veh.}&\rotatebox{90}{mov. person} &\rotatebox{90}{mov. bicyclist} &\rotatebox{90}{mov. motorcyc.} & \rotatebox{90}{\textbf{mIoU} $[$\%$]$}& \rotatebox{90}{\textbf{Time }$[$ms$]$}\\
		\midrule
        
        SpSequenceNet~\cite{shi2020cvpr} &88.5&24.0&26.2&29.2&22.7&6.3&0.0&0.0&90.1&57.6&73.9&27.1&91.2&66.8&84.0&66.0&65.7&50.8&48.7&53.2&0.1&2.3&26.2&41.2&36.2&43.1  & 497.3 \\
		\midrule
  
        TemporalLidarSeg~\cite{Duerr20203dv} &92.1&47.7&40.9&39.2&35.0&14.4&0.0&0.0&91.8&59.6&75.8&23.2&89.8&63.8&82.3&62.5&64.7&52.6&60.4& 68.2&2.1&12.4&40.4&42.8& 12.9&47.0 & -\\
		\midrule
        KPConv~\cite{thomas2019iccv}&93.7&44.9&47.2&42.5&38.6&21.6&0.0&0.0&86.5&58.4&70.5&26.7&90.8&64.5&84.6&70.3&66.0&57.0&53.9&69.4&5.8&4.7&67.5&67.4&47.2&51.2  & 139.6\\
		\midrule

        Cylinder3D~\cite{zhou2022tpami}&93.8&67.6&63.3&41.2&37.6&12.9&0.1&0.1&90.4&66.3&74.9&32.1&92.4&65.8&85.4&72.8&68.1&62.6&61.3&68.1&0.0&0.1&63.1&60.0&0.4&51.5 & 125.0\\
		\midrule
		MarS3D~\cite{liu2023cvpr}&95.1&49.2&49.5&39.7&36.6&16.2&\textbf{1.2}&0.0&89.9&66.8&74.3&26.4&92.1&68.2&86.0&72.1&70.5&62.8&64.8&78.4&5.1&10.0&58.0&67.3&36.3&52.7 & 180.5\\
		\midrule
        MemorySeg~\cite{Li2023iccv}&94.0&\textbf{68.3}&\textbf{68.8}&51.3&40.9&27.0&0.3&2.8&89.9&64.3&74.8&29.2&92.2&69.3&84.8&75.1&70.1&65.5&68.5&71.7&\textbf{13.6}&15.1&71.7&\textbf{74.4}&73.9&58.3& -\\
		\midrule

        SVQNet~\cite{Chen2023iccv}&\textbf{96.1}&64.4&60.3&40.4&\textbf{60.9}&\textbf{27.4}&0.0&0.0&\textbf{93.2}&\textbf{71.6}&\textbf{80.5}&\textbf{37.0}&\textbf{93.7}&\textbf{72.6}&\textbf{87.3}&\textbf{76.7}&\textbf{72.3}&\textbf{68.4}&\textbf{71.0}&80.5&3.9&7.5&\textbf{84.7}&72.4&\textbf{91.0}&60.5 & -\\
		\midrule

		Ours&95.1&60.7&59.1&\textbf{54.5}&55.3&17.8&\textbf{19.3}&\textbf{31.0}&91.6&64.9&76.4&35.2&93.2&71.6&85.7&72.5&70.1&62.7&69.5&\textbf{84.8}&11.2&\textbf{30.5}&60.6&72.5&76.3& \textbf{60.9}  & \textbf{67.1} \\
		\bottomrule
	\end{tabular}
 	\begin{tablenotes}
	\footnotesize
	\item - indicates the runtime is not available due to the baseline not being open-sourced. 
	\end{tablenotes}
	
	\label{tab:multi-scan}
\end{table*}

\subsection{Loss Function}
\label{sec:loss}
Due to multi-task setup, our loss function includes instance detection loss~$L_\text{det}$, MOS loss~$L_\text{mos}$, SSS loss~$L_\text{s\_sem}$ and multi-scan semantic segmentation loss~$L_\text{m\_sem}$.
In order to achieve a balance between the magnitudes of different loss functions and to expedite the convergence speed, we apply a weighted multi-task loss~\cite{liebel2018arxiv} $L_\text{total}$ to supervise the training, defined as:
\begin{equation}
	L_\text{total} =\sum_{i \in \{\text{det,mos,s\_sem,m\_sem}\}} \frac{1}{2 \sigma_{i}^2}L_i+ \ln (1+\sigma_{i}^2),
\end{equation}
where $\sigma_{i}$ is a learnable parameter used to represent the uncertainty of $L_i$.
$L_\text{det}$ is composed of instance classification loss and bounding box regression loss.  More details can be found in the CenterHead~\cite{yin2021cvpr}.
In addition, we employ the widely used weighed Cross-Entropy Loss function \cite{sun2022iros} for $L_\text{mos}$, $L_\text{s\_sem}$ and $L_\text{m\_sem}$:
\begin{equation}
	L_\text{\{mos,s\_sem,m\_sem\}}(y,\hat{y}) \hspace{-0.8mm}=\hspace{-0.8mm}  -\hspace{-1mm}\sum \alpha_i p(y_i)\text{log}(p(\hat{y}_i)), \alpha_i \hspace{-0.8mm}=\hspace{-0.8mm} 1/\sqrt{f_i}, 
\end{equation} 
where $y_i$ and $\hat{y}_i$ denote the ground truth and the predicted labels, respectively. 
$f_i$ is the frequency of the $i$-th class.

\section{Experimental Evaluation}
\label{sec:exp}

In this section, we conduct a series of experiments on SemanticKITTI~\cite{behley2019iccv} and nuScenes~\cite{Caesar2019cvpr} datasets to demonstrate SegNet4D's capabilities on multi-scan semantic segmentation (\secref{sec:mulit-scan}) and MOS (\secref{sec:mos}) and compare the performance with SOTA.
Subsequently, we integrate the method into a real unmanned ground platform (\secref{sec:nudt-campus}), supporting online real-time operation.
Besides, we perform ablation experiments to evaluate the effectiveness of the framework and the proposed modules (\secref{sec:Ablation}).
These experimental results will substantiate our claims regarding the contributions.

\textbf{Datasets-SemanticKITTI~\cite{behley2019iccv}.} 
SemanticKITTI provides semantic labels for each individual LiDAR scan in the KITTI~\cite{Geiger2012cvpr} odometry dataset, which comprises 22 sequences collected with a Velodyne HDL-64E LiDAR.
Following previous work~\cite{zhou2022tpami,Li2023iccv}, we use the standard data split where sequences 00 to 10 are used for training (with sequence 08 for validation), and sequence 11 to 21 for testing. 
The dataset contains multiple semantic category labels.
In our model, the training is supervised with 26 semantic categories, including 6 moving classes, 19 static classes, and one $outlier$ class.
For the MOS task, All the semantic categories are reorganized into two classes: moving and static~\cite{chen2021ral,mersch2022ral}.
Due to the imbalanced distribution of moving objects between the training and test set, an additional dataset, KITTI-Road, is introduced in \cite{sun2022iros} to mitigate the impact.
We follow the experimental setups described in~\cite{sun2022iros} to test the MOS performance.

\textbf{Datasets-nuScenes~\cite{Caesar2019cvpr}.}
It consists of 1000 driving scenes collected with a 32-beam LiDAR sensor.
For point cloud segmentation tasks, it is mainly used for evaluating SSS with 16 static semantic classes.
To achieve multi-scan segmentation, we utilize the annotated bounding box motion attributes to generate 8 new moving   categories, including $moving \; car$, $moving \; bus$, $moving \;truck$, $moving \;construction \;vehicle$, $moving\; trailer$, $moving\; motorcyclist$, $moving \;bicyclist$, and $moving \;person$.
We will release the multi-scan semantic segmentation dataset of nuScenes for the convenience of the community.
Similar to the SemanticKITTI, we divide all semantic categories into two classes (moving and static) for evaluating MOS.

\textbf{Implementation Details}.
We restrict the point cloud range in $[x:\pm 60m,\;y:\pm50m,\;z:-4m\sim2m]$ for SemanticKITTI, $[x:\pm 50m,\;y:\pm50m,\;z:-4m\sim2m]$ for nuScenes, and set $g=0.1m$ for encoding motion features.
For multi-scan semantic segmentation, we set $C=20,C'=26$ for SemanticKITTI and $C=17,C'=25$ for nuScenes.
The proposed model is built on the PyTorch~\cite{paszke2019neurips} library and trained with 4 NVIDIA RTX 3090 GPUs. 
We set the batch size to 8 on a single GPU and train the network for a total of 80 epochs.
The learning rate is initialized as $10^{-4}$ in the Adam optimizer~\cite{kingma2014iclr} and a decay factor of 0.01 for each epoch.
During the training process, we employ widely-used data augmentation techniques such as random flipping, scaling, and rotation to improve model performance. 

\textbf{Evaluation Metrics.} For MOS performance evaluation, we use the Intersection-over-Union (IoU) \cite{everingham2010ijcv} of moving objects as the metric:
\begin{equation}
\text{IoU} = \frac{\text{TP}}{\text{TP}+\text{FP}+\text{FN}},
\end{equation}
where TP, FP, and FN represent the predictions of the moving class that are classified as true positive, false positive, and false negative, respectively.
For multi-scan semantic segmentation, we use the mean Intersection-over-Union (mIoU) across all categories as the evaluation metric.

\begin{table*}[t]
	\fontsize{7pt}{7pt}\selectfont
        \setlength{\tabcolsep}{2.5pt}
	\caption{Multi-scan semantic segmentation performance evaluation on the nuScenes dataset. “mov." denotes moving category. }
	\centering
    \begin{tabular}{l|cccccccccccccccccccccccc|p{0.65cm}<{\centering}}
        \toprule
        Methods & \rotatebox{90}{barrier} &\rotatebox{90}{bicycle}&\rotatebox{90}{bus} &\rotatebox{90}{car} &\rotatebox{90}{construction} & \rotatebox{90}{motorcycle}& \rotatebox{90}{pedestrian}& \rotatebox{90}{traffic cone}&\rotatebox{90}{trailer} & \rotatebox{90}{truck}& \rotatebox{90}{driveable}&\rotatebox{90}{other flat} & \rotatebox{90}{sidewalk}& \rotatebox{90}{terrain}& \rotatebox{90}{manmade}& \rotatebox{90}{vegetation}& \rotatebox{90}{mov. car}& \rotatebox{90}{mov. bus}&\rotatebox{90}{mov. truck}& \rotatebox{90}{mov. const.}& \rotatebox{90}{mov. trailer}& \rotatebox{90}{mov. motor.}&\rotatebox{90}{mov. bicyc.} &\rotatebox{90}{mov. person}  & \rotatebox{90}{\textbf{mIoU} $[$\%$]$}  \\
        \midrule
        SpSequenceNet~\cite{shi2020cvpr}&65.3&0.0&0.0&0.0&23.2&0.0&52.2&39.2&26.3&52.6&94.5&67.5&67.6&70.5&82.1&81.3&46.0&33.4&26.0&0.0&18.9&38.9&0.0&52.5&39.1\\
        \midrule
        KPConv~\cite{thomas2019iccv}&65.0&32.0&58.3&59.3&39.4&37.4&27.6&49.3&29.9&61.8&79.3&55.7&60.0&55.5&44.6&41.8&54.3&49.5&47.9&0.0&\textbf{42.3}&68.9&39.4&57.6&48.2\\
        \midrule
        Cylinder3D~\cite{zhou2022tpami}&70.0&22.3&41.6&74.0&40.8&22.3&43.3&59.3&\textbf{46.1}&62.3&95.2&64.8&70.2&71.3&\textbf{87.7}&86.0&56.6&43.5&40.0&0.0&30.3&65.3&22.3&63.9&53.3\\
        \midrule
        MarS3D~\cite{liu2023cvpr}&70.5&24.7&60.0&\textbf{79.9}&32.0&34.9&\textbf{51.3}&53.0&10.4&66.0&95.4&59.9&72.7&\textbf{75.8}&87.2&\textbf{86.1}&66.5&48.0&52.4&0.0&23.1&69.0&9.7&\textbf{72.7}&54.3\\
        \midrule
        Ours&\textbf{77.4}&\textbf{32.6}&\textbf{63.8}&73.8&\textbf{41.1}&\textbf{44.0}&51.2&\textbf{63.2}&42.3&\textbf{74.2}&\textbf{96.2}&\textbf{69.3}&\textbf{74.2}&73.5&64.6&55.9&\textbf{68.6}&\textbf{51.3}&\textbf{59.4}&0.0&27.2&\textbf{74.3}&\textbf{40.8}&72.4& \textbf{57.9} \\
        \bottomrule
	\end{tabular}
	\vspace{-0.1cm}
	\label{tab:multi-scan_nuscenes}
\end{table*}

\begin{figure*}[ht]
	\centering
	\includegraphics[width=1\linewidth]{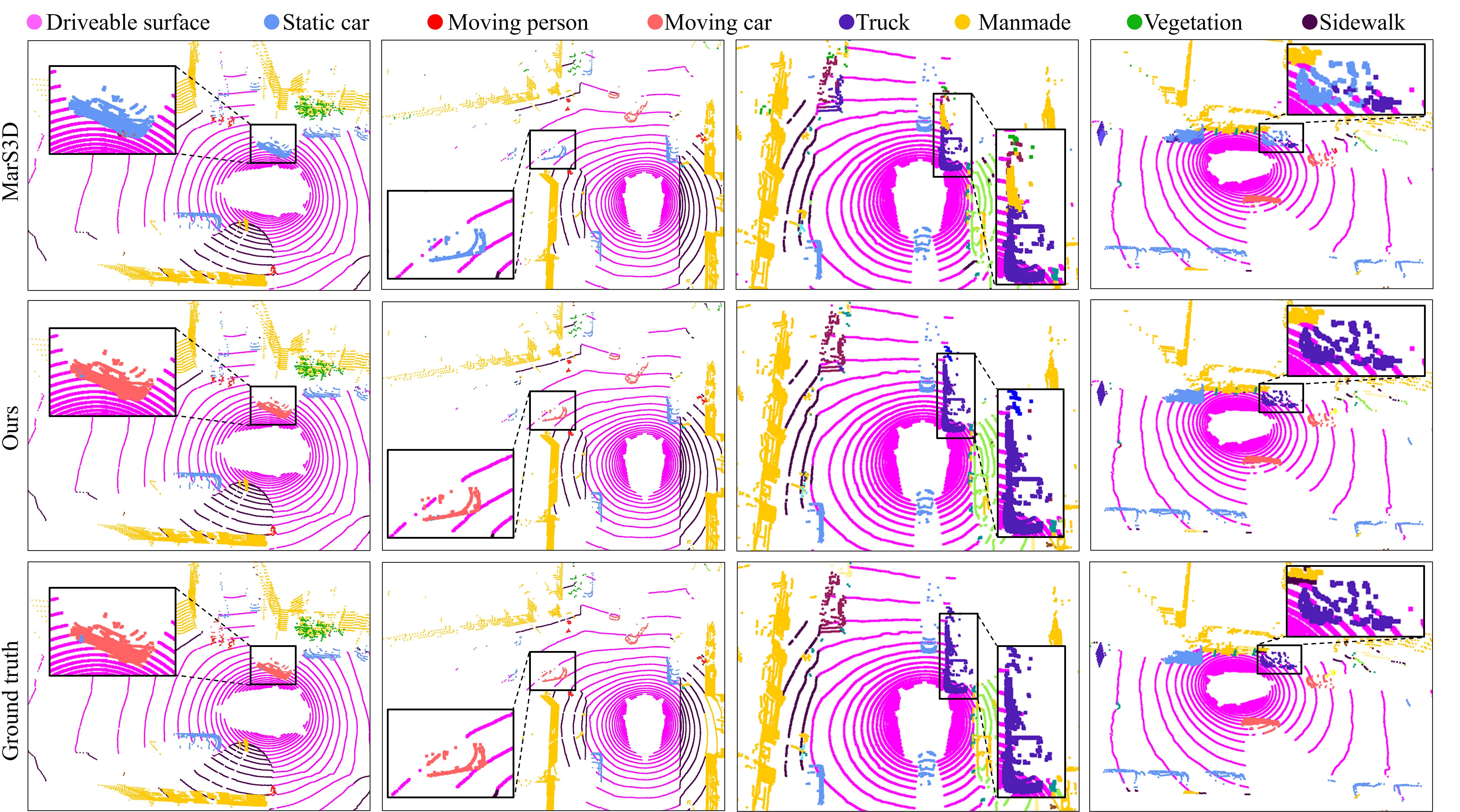}
	\caption{The qualitative results comparison between our approach and MarS3D on the nuScenes dataset.
	}
	\label{fig:compare_sem}
    \vspace{-0.2cm}
\end{figure*}

\subsection{Evaluation for Multi-scan Semantic Segmentation}
\label{sec:mulit-scan}
We evaluate the 4D semantic segmentation performance of our approach on the SemanticKITTI multi-scan semantic segmentation benchmark and nuScenes validation dataset, and compare the results with LiDAR-only SOTA baselines, including (a) single-scan-based methods (stack historical LiDAR scans into a single pointcloud as input for multi-scans semantic segmentation): 
KPConv\cite{thomas2019iccv}, 
Cylinder3D\cite{zhou2022tpami}; and specially designed multi-scans semantic segmentation \mbox{methods}: SpSequenceNet~\cite{shi2020cvpr}, 
TemporalLidarSeg~\cite{Duerr20203dv}, 
MarS3D\cite{liu2023cvpr}, SVQNet~\cite{Chen2023iccv} and MemorySeg~\cite{Li2023iccv}.
Like all methods, we only utilize the past two LiDAR scans to predict semantic labels for a fair comparison, i.e., $N=3$.

The quantitative results are presented in~\tabref{tab:multi-scan}. Our approach achieves a mIoU of 60.9\% on the SemanticKITTI dataset and outperforms all methods, demonstrating its effectiveness.
For semantic class $bicyclist$ and $motorcyclist$, our SegNet4D obtains a significant improvement, indicating that the incorporation of instance information enables our model to identify these foreground points more effectively.
Besides, we also test the average time for all open-source methods on sequence 08 using a single NVIDIA RTX 3090 GPU.
From the results, the existing methods all have a runtime exceeding 100ms, rendering them unsuitable for real-time operation.
In contrast, our method achieves a runtime of 67.1ms and is the only approach capable of real-time operation, showing that our method is highly efficient for 4D LiDAR semantic segmentation.

Additionally, we further evaluate our approach on the nuScenes datasets and compare the performance with other baselines.
As shown in~\tabref{tab:multi-scan_nuscenes}, our method consistently delivers the best performance, demonstrating its adaptability to different types of LiDAR and scene changes.
The qualitative comparison with MarS3D, as illustrated in~\figref{fig:compare_sem}, reveals our approach's superior ability for moving object recognition. This is also further corroborated by quantitative analyses presented in the~\secref{sec:mos}.
Additionally, our method exhibits complete segmentation for big instances, like trucks, contrasting with Mars3D's incomplete segmentation results. This contrast highlights the effectiveness of our instance-aware design.


\subsection{Evaluation for Moving Object Segmentation}
\label{sec:mos}
We evaluate the result on the SemanticKITTI-MOS benchmark and nuScenes dataset, and compare it with SOTA MOS methods, including (a) projection-based: LMNet~\cite{chen2021ral}, MotionSeg3D~\cite{sun2022iros}, RVMOS~\cite{kim2022ral},  MotionBEV~\cite{zhou2023ral} and MF-MOS~\cite{cheng2024icra}; (b) point-based: 4DMOS~\cite{mersch2022ral} and InsMOS~\cite{wang2023iros}; as well as multi-scan segmentation baselines:  KPConv~\cite{thomas2019iccv}, SpSequenceNet~\cite{shi2020cvpr}, Cylinder3D~\cite{zhou2022tpami} and MarS3D~\cite{liu2023cvpr}.

The quantitative comparison is presented in~\tabref{tab:performance_test}. 
Our method achieves the best results on both the SemanticKITTI and nuScenes dataset, demonstrating its superior performance in motion segmentation.
SegNet4D also shows improved performance compared to original InsMOS~\cite{wang2023iros}, indicating that semantic information is beneficial for the identification of moving object, as it provides a vital cue for distinguishing between movable and immovable classes. 
Additionally, our method significantly outperforms other multi-scan semantic segmentation methods in terms of MOS performance, benefiting from our framework's explicit MOS supervision during the training phase.

\begin{table}[t]
	\footnotesize
        \renewcommand\arraystretch{1.1}
        \setlength{\tabcolsep}{9pt}
	\caption{Moving object segmentation performance evaluation. }
	\centering
	\begin{threeparttable}
	\begin{tabular}{lp{1.5cm}<{\centering}cc}
		\toprule
		 Methods               & Publication &  \makecell[c]{Semantic\\KITTI}  & nuScenes   \\
		\midrule
		LMNet~\cite{chen2021ral}                   & RAL 2021 &   62.5 &   49.9 \\
		4DMOS~\cite{mersch2022ral}                   & RAL 2022 &   65.2 &   67.8 \\
		MotionSeg3D~\cite{sun2022iros}         & IROS 2022 &   70.2 &   63.0 \\
        RVMOS~\cite{kim2022ral}                   & RAL 2022 &   74.7 &   - \\
		InsMOS~\cite{wang2023iros}             & IROS 2023 &   75.6 &   65.7 \\
        MotionBEV~\cite{zhou2023ral}               & RAL 2023 &   75.8 &   64.7 \\
        MFMOS~\cite{cheng2024icra}                   & ICRA 2024 &   76.7 &   62.6 \\
		\midrule
        KPConv~\cite{thomas2019iccv}                  & ICCV 2019 &  60.9 &  56.2 \\
        SpSequenceNet~\cite{shi2020cvpr}           & CVPR 2020 &  43.2 &  42.9\\
        Cylinder3D~\cite{zhou2022tpami}              & TPAMI 2022 &  61.2 &  59.5\\
        MarS3D~\cite{liu2023cvpr}                  & CVPR 2023 &  66.2 &  64.3\\
		Ours              & - & \textbf{78.5} & \textbf{69.0}\\
		\bottomrule
	\end{tabular}

	\end{threeparttable}
	\label{tab:performance_test}
\end{table}

\subsection{Operation on Real-World Platform}
\label{sec:nudt-campus}
\begin{figure}[t]
	\centering
	\includegraphics[width=0.98\linewidth]{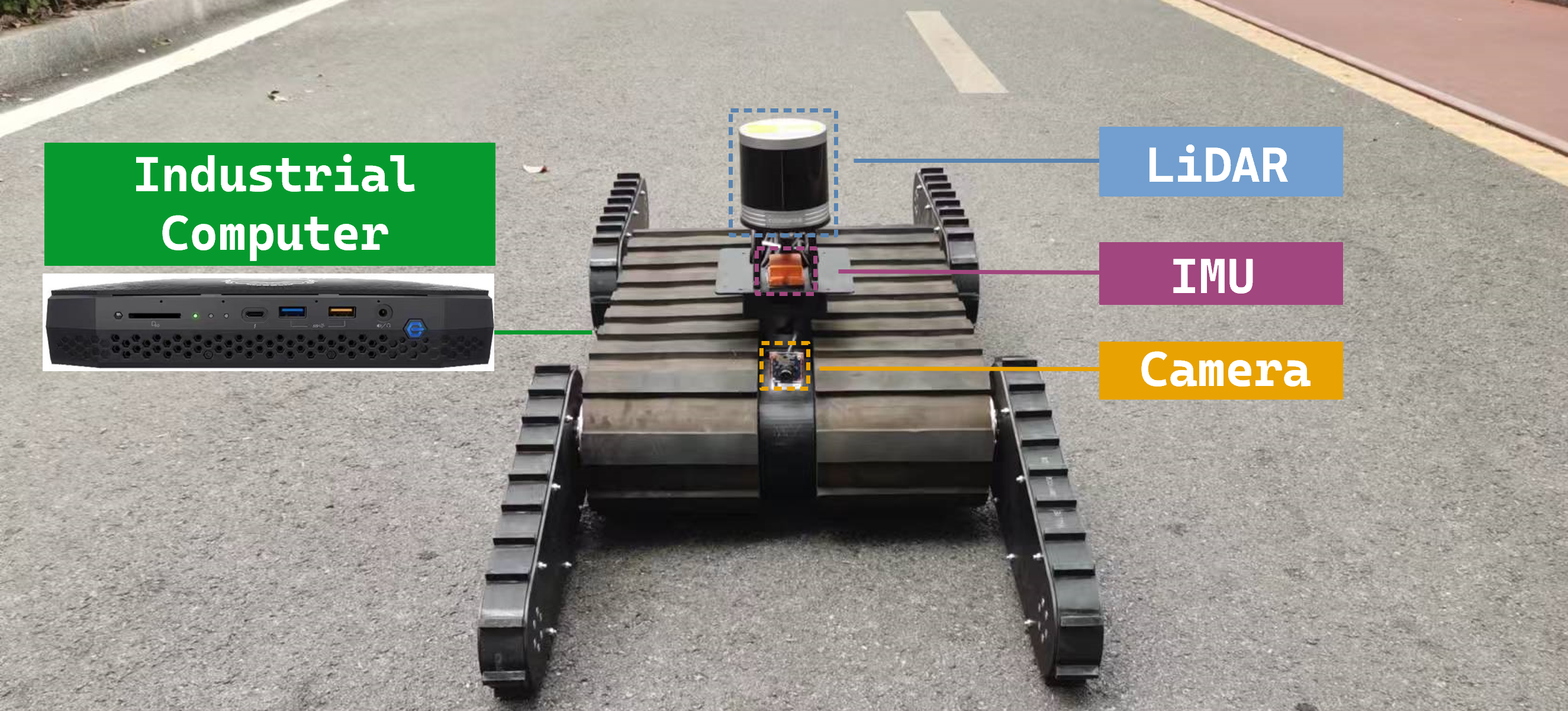}
	\caption{Self-developed unmanned ground platform for experiments.}
	\label{fig:robot}
    \vspace{-0.3cm}
\end{figure}

To demonstrate the real-world utility of SegNet4D, we further simplify our model and integrate it into a unmanned ground system.
As shown in~\figref{fig:robot}, our platform is equipped with a RoboSense Helios-32 LiDAR with 10~Hz sampling frequency, an Xsens MTI-300 IMU, a CMOS camera, and an industrial computer with Intel Core i7-1165G7, 16G RAM, and NVIDIA RTX 2060 GPU.

\begin{figure}[t]
	\centering
	\includegraphics[width=0.98\linewidth]{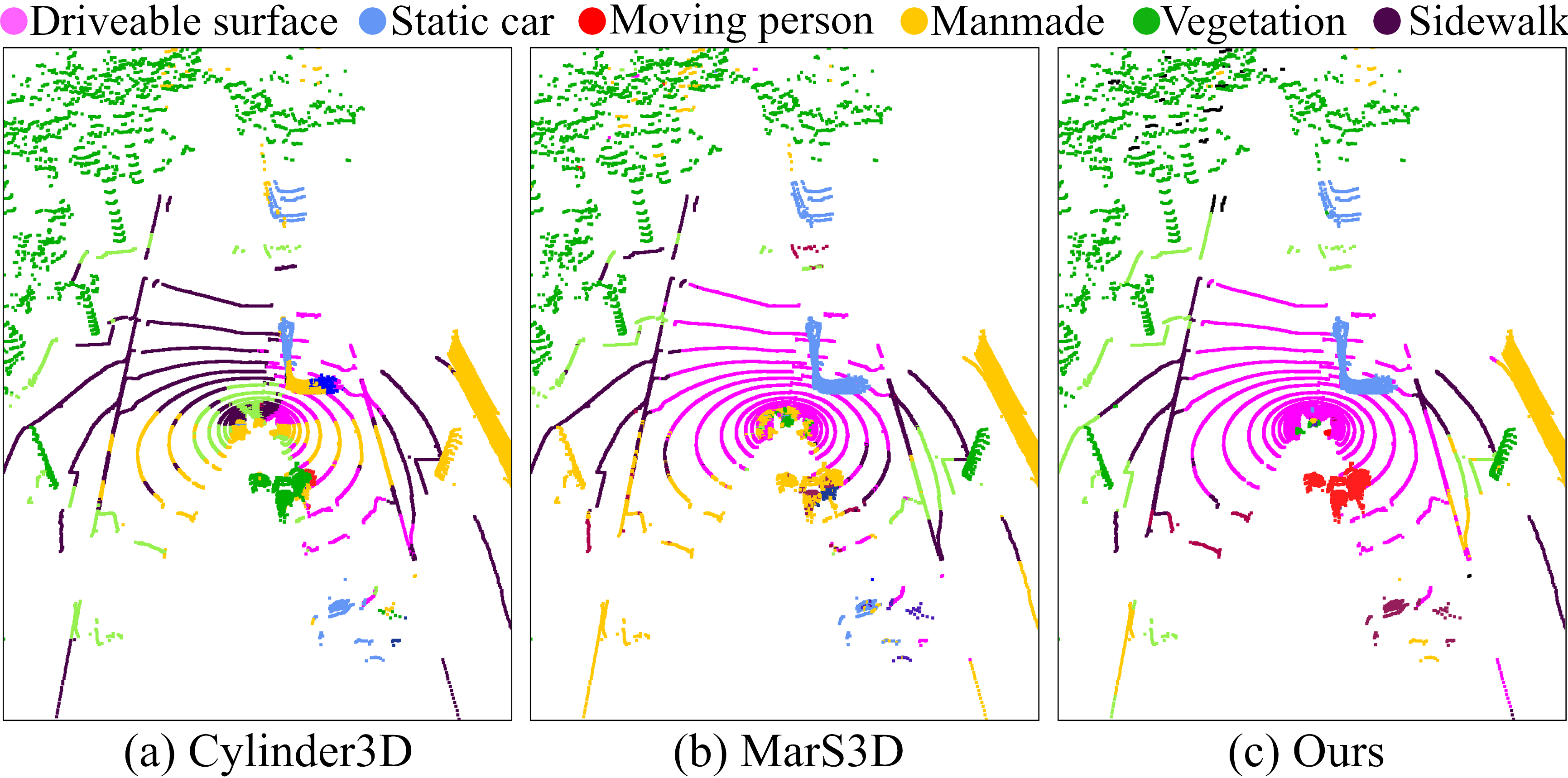}
	\caption{The qualitative comparison on our campus scene.}
	\label{fig:nudt-campus}
 \vspace{-0.3cm}
\end{figure}

Our method achieves real-time operation at 15.7\,Hz rate on the self-developed platform, faster than the typical rotation LiDAR sensor frame-rate of 10\,Hz. Its parameter count is merely 35.7\,M, indicating small memory consumption suitable for limited onboard resources.
We also utilize this platform to collect a dataset from the campus scene and qualitatively compare the semantic segmentation results with Cylinder3D~\cite{zhou2022tpami} and Mars3D~\cite{liu2023cvpr}. 
All methods are only trained on the nuScenes dataset and directly applied for inference on our dataset. 
As illustrated in~\figref{fig:nudt-campus}, we can see that our SegNet4D gets the best segmentation results, demonstrating its practical utility for the real platform.
Additionally, we manually annotate a sequence (900 frames) with moving labels to quantitatively evaluate the MOS generalization performance. 
From the results shown in~\tabref{tab:nudt-campus}, our method still possesses the best generalization ability.

The above-mentioned experimental results show that SegNet4D can operate in real-time on the real unmanned ground platform while exhibiting superior performance in 4D semantic segmentation and MOS. This highlights its practical utility in enhancing robots' environmental perception capabilities.

\begin{table}[t]
	\footnotesize
        \renewcommand\arraystretch{1.1}
        \setlength{\tabcolsep}{9pt}
	\caption{MOS generalizability performance evaluation on our campus dataset. The best results are in bold.}
	\centering
	\begin{threeparttable}
	\begin{tabular}{lp{3.5cm}<{\centering}c}
		\toprule
		 Methods & &  IoU$[$\%$]$   \\
		\midrule
        Cylinder3D~\cite{zhou2022tpami} &  &  25.2 \\
        MarS3D~\cite{liu2023cvpr}    &  &  40.4 \\
        \midrule
		Ours      &    &    \textbf{49.8} \\
		\bottomrule
	\end{tabular}
	\end{threeparttable}
	\label{tab:nudt-campus}
\end{table}

\subsection{Ablation Study}
\label{sec:Ablation}

In this section, we conduct a series of ablation experiments on our framework and MSFM module to test their effectiveness.
The results are shown in~\tabref{tab:ablation_network}, 
Here, “Instance" indicates whether instance information is integrated into the Upsample Fusion Module to achieve instance-aware segmentation. 
“One Head” means directly predicting multi-scan semantic labels using only one head, indicating an end-to-end fashion.
“Two Head" means utilizing two heads (the motion head and the semantic head, i.e., our framework) to predict moving objects and single-scan semantic labels, respectively, and finally merging their results to achieve multi-scan semantic segmentation.
“Refinement" refers to moving instance refinement mentioned in Section~\ref{sec:Refinement}, and it only focuses on refining the motion predictions.

By comparing $[$A$]$ and $[$C$]$, we can see that incorporating instance information can improve the accuracy of MOS and semantic segmentation, which equips the network with instance-aware segmentation capabilities, and the results substantiate our second claim.
In the $[$B$]$ and $[$C$]$, predicting multi-scan semantic labels directly from a head is even inferior to the results obtained by manually integrating outputs from the motion head and the semantic head. 
This illustrates the superiority of our framework in comparison to directly predicting multi-scan semantic labels using an end-to-end manner, thereby supporting our first claims.
The results of $[$C$]$ and $[$D$]$ indicate that the  refinement is effective for improving the motion predictions, which further proves the utility of instance information.
Besides, we manually merge the predictions of MOS and SSS, labeled as Manual MS in the~\tabref{tab:ablation_network}, and compare its results with the output of the proposed MSFM.
We can see that the MSFM's segmentation result always outperforms manual fusion, confirming its effectiveness in integrating motion and static semantic predictions.


\begin{table}[t]
	\footnotesize
        \renewcommand\arraystretch{1.1}
        \setlength{\tabcolsep}{4pt}
	\caption{Ablation studies on the nuScenes validation set. \textbf{MOS}: moving object segmentation. \textbf{SSS}: single-scan semantic segmentation. \textbf{Manual MS}: manually merging MOS and SSS for multi-scan semantic segmentation. \textbf{Network MS}: the Network's predictions for multi-scan semantic segmentation, i.e., the output of MSFM for $[$A$]$,$[$C$]$,$[$D$]$, the output by One Head for $[$B$]$. }
	\centering
	\begin{threeparttable}
		\begin{tabular}{l|p{0.6cm}<{\centering}p{0.6cm}<{\centering}p{0.6cm}<{\centering}p{0.8cm}<{\centering}|p{0.6cm}<{\centering}p{0.6cm}<{\centering}p{0.9cm}<{\centering}p{1cm}<{\centering}}
			\toprule
			   &\makecell[c]{Inst-\\ance}  &   \makecell[c]{One\\Head}  & \makecell[c]{Two\\Head} & \makecell[c]{Refin-\\ement}&  MOS &  \makecell[c]{SSS} & \makecell[c]{Manual\\MS}  & \makecell[c]{Network\\MS} \\
			\midrule
			$[$A$]$&    &    & \Checkmark &  &  67.1  & 67.3  & 55.5  & 56.3\\
            $[$B$]$& \Checkmark  & \Checkmark  &   & &  -  & -  & - & 56.0 \\
            $[$C$]$&   \Checkmark &    &\Checkmark & &  68.5  & 68.6  & 57.4  & 57.9 \\
            $[$D$]$&  \Checkmark &    & \Checkmark & \Checkmark  &  69.0  & 68.6  & 57.5  & 57.9\\
			\bottomrule
		\end{tabular}
	\end{threeparttable}
	\label{tab:ablation_network}
\end{table}

\section{Conclusion}
\label{sec:conclusion}
In this paper, we present a novel 4D semantic segmentation method to predict both point-wise moving labels and semantic labels for LiDAR measurements, and operate in real-time.
The framework decomposes the complex 4D semantic segmentation task into MOS and SSS  tasks, finally merging their predictions to achieve more accurate 4D semantic segmentation.
We adopt a projection-based approach to quickly obtain motion features, which significantly reduces the computational complexity compared to 4D convolutions.
To achieve instance-aware segmentation, we concatenate the motion features with the spatial features of the current scan, feeding them into the network for instance detection, and subsequently inject the instance features into the prediction pipeline.
In addition, we design a motion-semantic fusion module to integrate the point-wise motion states and static semantic predictions, enabling motion-guided 4D semantic segmentation.
Extensive experiments on multiple datasets and a real-world unmanned ground platform demonstrate that our approach is effective and efficient.

\footnotesize{
\bibliographystyle{IEEEtran}
\bibliography{new,glorified}
}

\end{document}